\documentclass{article} %
\usepackage{iclr2026_conference,times}

\usepackage{amsmath,amsfonts,bm}

\def\eqref#1{equation~\ref{#1}}

\def\1{\bm{1}}

\DeclareMathAlphabet{\mathsfit}{\encodingdefault}{\sfdefault}{m}{sl}
\SetMathAlphabet{\mathsfit}{bold}{\encodingdefault}{\sfdefault}{bx}{n}

\usepackage{hyperref}
\usepackage{url}
\usepackage{graphicx}
\usepackage{booktabs}

\usepackage{amsmath} 
\usepackage{adjustbox}

\usepackage{xcolor}
\hypersetup{
    colorlinks=true,
    linkcolor=black,
    citecolor=black,
}

\usepackage{wrapfig}

\title{Expert Merging: Model Merging with Unsupervised Expert Alignment and Importance-Guided Layer Chunking}

\author{\textbf{Dengming Zhang}$^{1,2}$\thanks{Equal contribution.} \quad
        \textbf{Xiaowen Ma}$^{2}$\footnotemark[1] \quad
        \textbf{Zhenliang Ni}$^{2}$ \\ 
        \textbf{Zhenkai Wu}$^{1,2}$ \quad
        \textbf{Han Shu}$^{2}$ \quad
        \textbf{Xin Jiang}$^{2}$ \quad
        \textbf{Xinghao Chen}$^{2}$\thanks{Corresponding author: xinghao.chen@huawei.com} \\
        $^1$ Zhejiang University \quad
        $^2$ Huawei Noah's Ark Lab
}

\iclrfinalcopy %
\begin{document}

\maketitle
\begin{abstract}
Model merging, which combines multiple domain-specialized experts into a single model, offers a practical path to endow Large Language Models (LLMs) and Multimodal Large Language Models (MLLMs) with broad capabilities without the cost of joint training or serving many models. However, training-free methods rely on hand-tuned coefficients, whereas training-based methods primarily align parameters rather than downstream task behavior and typically treat all layers uniformly, ignoring inter-layer heterogeneity. We introduce Expert Merging, a training-light method that learns a small set of layer-wise coefficients using only unlabeled calibration data. The coefficients are optimized to explicitly align the merged model’s hidden states and logits with those of the corresponding experts, with a coefficient regularizer for stability and task-weighted losses for controllable trade-offs. To capture inter-layer variation, Expert Merging++ augments this design with importance-guided chunking: a normalized layer-importance metric, derived from learned coefficients, task-vector magnitudes, and parameter counts, allocates more chunk-wise coefficients to high-importance layers while keeping low-importance layers lightweight. The result is a label-free, parameter-efficient, and scalable approach to multi-expert model merging across LLMs and MLLMs. Across MLLM backbones (InternVL and Qwen2-VL) and the LLM backbone (Mistral), our method surpasses strong training-free and training-based merging baselines, with Expert Merging++ delivering further gains and, in some cases, even exceeding supervised Mixture Training. The source code is available at \url{https://github.com/Littleor/ExpertMerging}

\end{abstract}

\section{Introduction}
\label{sec:intro}

Large language models (LLMs) show strong general abilities but often lag on domain-specific tasks that require specialized knowledge~\citep{hadi2023survey, ling2023domain}. A standard remedy is Supervised Fine-Tuning (SFT) that turns a base model into domain-specific expert models for code, math, or reasoning~\citep{zhao2023survey, dodge2020fine, chung2024scaling}. However, maintaining and serving a separate model per domain imposes substantial storage, memory, and engineering overhead, hindering deployment at scale~\citep{yadav2024matters}. Model merging offers an attractive alternative: combine multiple domain experts into a single multi-competent model while retaining their respective strengths~\citep{yu2024language,wortsman2022model,ilharco2022editing}. Existing model merging methods broadly fall into two groups: training-free methods with predefined coefficients and training-based methods that learn coefficients or parameters~\citep{wei2025unifying}.

Among training-free approaches, Task Arithmetic~\citep{ilharco2022editing} as a classic method forms a merged model by adding expert task vectors (i.e., subtracting the base model parameters from the SFT model parameters) to the base model with fixed, predefined coefficients. Building on this idea, subsequent training-free approaches refine the merging process to mitigate parameter redundancy and interference, reporting measurable improvements~\citep{yu2024language,gargiulo2025task}. However, these methods typically rely on hand-specified hyperparameters (e.g., per-task coefficients), which are selected via manual tuning or grid search. More importantly, these methods often focus solely on parameter-space alignment while neglecting task alignment.

Training-based methods address these limitations by learning optimal merging parameters or coefficients through gradient-based optimization. WUDI~\citep{cheng2025whoever} reduces cross-task interference by minimizing a layer-wise interference objective, improving the performance of the merged model. Building on this idea, WUDI v2~\citep{wei2025unifying} performs singular value decomposition (SVD) before training to isolate task-salient signals from noise, further enhancing model performance. Nevertheless, these approaches remain primarily parameter-space alignment procedures and overlook the merged model’s downstream performance. 
AdaMerging~\citep{AdaMerging_ICLR_2024} optimizes task-wise or layer-wise coefficients by minimizing prediction entropy to obtain a favorable combination of coefficients. While this accounts for task performance, entropy minimization provides no explicit task alignment signal and can induce overconfidence under distribution shift; moreover, assigning a single trainable coefficient per layer ignores inter-layer heterogeneity, ultimately constraining performance.

To address both task alignment and inter-layer heterogeneity, we introduce Expert Merging and its extension, Expert Merging++. For task alignment, Expert Merging learns trainable layer-wise coefficients that align the merged model’s hidden states and logits with those of the corresponding expert models on a small set of unlabeled inputs (5–10 samples), thereby explicitly matching downstream behavior. To mitigate distribution shifts during merging, we incorporate a coefficient regularization term that stabilizes optimization. For inter-layer heterogeneity, our analysis reveals substantial variation across layers in optimal coefficient magnitudes and parameter number. 
We therefore propose a layer-importance metric to quantify each layer’s contribution and a chunk-wise coefficient method: high-importance layers are partitioned into structured parameter chunks with distinct coefficients, while low-importance layers remain lightweight.
We evaluate these methods on LLMs and multimodal large language models (MLLMs): Expert Merging consistently outperforms strong training-free and prior training-based baselines in both settings, Expert Merging++ yields additional gains, and on MLLMs the approach can even surpass supervised Mixture Training.

Our contributions are: (1) Expert Merging that learns layer-wise coefficients to align the merged model’s hidden states and logits to expert models on unlabeled data, with coefficient regularization for stability; (2) Expert Merging++ with chunk-wise coefficients guided by a layer-importance metric to allocate more capacity by chunking high-importance layers while keeping low-importance layers lightweight; and (3) comprehensive results and ablations on both an LLM (Mistral-7B) and MLLMs (InternVL, Qwen2-VL) showing Expert Merging surpasses training-free and prior training-based baselines, with Expert Merging++ yielding further gains.

\section{Related Work}
Model merging aims to combine multiple task-specific models into a unified model with diverse capabilities~\citep{wortsman2022model}. This approach has gained significant attention in the large model domain, particularly for large language models~\citep{wei2025unifying, zhou2024metagpt, goddard-etal-2024-arcees}, due to not requiring access to original training datasets and its cost-effective nature~\citep{stoica2023zipit}. Existing model merging approaches can be categorized into two main classes based on whether they require training: training-free methods and training-based methods~\citep{wei2025unifying}. 

\subsection{Training-Free Model Merging Methods}

Training-free methods~\citep{wortsman2022model, gargiulo2025task, marczak2025no} operate directly on model parameters without requiring additional optimization procedures. These approaches typically manipulate pre-trained model weights through mathematical operations to achieve effective model combination. Task Arithmetic~\citep{ilharco2022editing} represents a foundational approach in this category, computing task vectors for each task-specific model by subtracting the base model parameters from fine-tuned parameters. The method then merges multiple task-specific models through predefined scaling coefficients. However, this approach often suffers from parameter interference when combining multiple tasks. TIES Merging~\citep{yadav2023ties} addresses the parameter conflict problem inherent in Task Arithmetic by proposing a sophisticated conflict resolution mechanism. The method analyzes parameter changes based on both magnitude and direction, trimming conflicting parameters before merging. Specifically, TIES identifies parameters with conflicting signs across different task vectors and resolves these conflicts through magnitude-based voting schemes. DARE~\citep{yu2024language} introduces a different perspective by focusing on parameter sparsification during the merging process. The method applies drop and rescale operations to task vectors, randomly dropping a fraction of parameters and rescaling the remaining ones to maintain the expected magnitude. This approach reduces parameter interference by randomly dropping parameters, demonstrating that sparsification can actually improve merging performance.
Despite their simplicity, training-free approaches typically rely on fixed, hand-tuned coefficients and operate purely in parameter space without explicit task-level alignment, making them sensitive to hyperparameters and prone to degraded performance when many experts are combined.

\subsection{Training-Based Model Merging Methods}

Training-based methods optimize model combination through gradient-based updates, enabling more adaptive merging strategies. These methods can be further categorized by their data requirements.
Sample-independent methods eliminate the dependency on training samples while still employing optimization. WUDI~\citep{cheng2025whoever} reduces cross-task interference by minimizing a layer-wise interference objective, aligning the merged model’s activations with those of the experts and improving robustness. Building on this idea, WUDI v2~\citep{wei2025unifying} performs a singular value decomposition (SVD) of task vectors before training to isolate task-salient signals from noise, further improving performance.
Sample-dependent methods leverage data to optimize merging parameters. AdaMerging~\citep{AdaMerging_ICLR_2024} optimizes task-wise or layer-wise coefficients by minimizing prediction entropy to obtain a favorable combination of coefficients. While effective, entropy minimization provides no explicit task-alignment signal and can induce overconfidence under distribution shifts; moreover, assigning a single trainable coefficient per layer overlooks inter-layer heterogeneity, which can limit performance.

Despite progress, most training-based methods emphasize parameter-space alignment or implicit distribution matching, provide limited control over task trade-offs, and ignore layer heterogeneity. We instead align hidden states/logits on unlabeled data with regularized layer-wise and importance-guided chunk-wise coefficients, improving alignment, trade-off control, and performance.

\section{Method}
\label{sec:method}

\begin{figure}[t]
\begin{center}
\includegraphics[width=\textwidth]{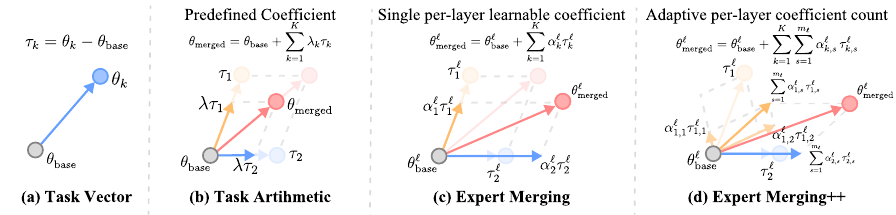}
\end{center}
\caption{Conceptual overview: (a) Task Vector, (b) Task Arithmetic (TA), (c) Expert Merging, and (d) Expert Merging++ with importance-guided chunk-wise coefficients.}
\label{fig:vector}
\end{figure}

We propose Expert Merging, a training-light method that merges multiple domain experts via layer-wise coefficients learned from unlabeled data to align hidden states and logits; Expert Merging++ further allocates capacity by layer importance with chunk-wise coefficients (Figure~\ref{fig:vector}).

\subsection{Preliminaries}
\label{subsec:prelim}

Let $\theta_{\text{base}}$ denote the parameters of a base model $f(\cdot;\theta_{\text{base}})$ with $L$ layers, and let $\{\theta_{k}\}_{k=1}^K$ be the parameters of $K$ domain experts obtained by SFT on tasks $\{\mathcal{T}_k\}_{k=1}^K$. Define the task vector of expert $k$ as $\tau_k = \theta_{k} - \theta_{\text{base}}$ and its layer-$\ell$ block as $\tau^{\ell}_k$. Training-free Task Arithmetic forms a merged model by $\theta_{\text{TA}}=\theta_{\text{base}}+\sum_{k=1}^K \lambda_k\,\tau_k$ with predefined coefficients $\{\lambda_k\}$ (panel (b) of Figure~\ref{fig:vector}). In contrast, we seek \emph{trainable, layer-wise/chunk-wise} coefficients that are learned from unlabeled data.

We assume access to an unlabeled, task-partitioned calibration set $\mathcal{D}=\bigcup_{k=1}^K \mathcal{D}_k$ where $\mathcal{D}_k$ contains inputs reflecting the distribution of task $\mathcal{T}_k$ (text-only for LLMs; image+text for MLLMs). From $\mathcal{D}$ we learn \emph{layer-wise} (and later \emph{chunk-wise}) coefficients, with the objective of aligning the merged model's performance with expert $k$ on inputs $x\in\mathcal{D}_k$.

\subsection{Expert Merging}
\label{subsec:expert-merging}

\begin{figure}[t]
\centering
\includegraphics[width=\linewidth]{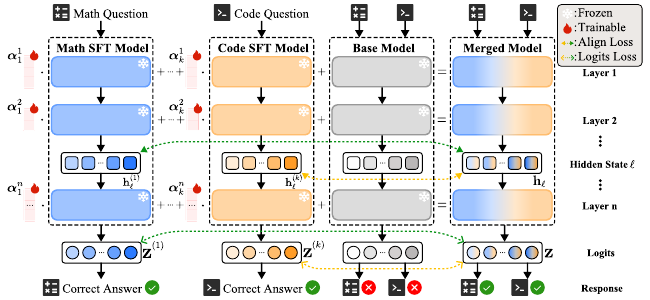}
\caption{Architecture of Expert Merging++. The base and expert models are frozen (snowflakes) and only the chunk-wise coefficients $\{\boldsymbol{\alpha}_k^{\ell}\}$ are trainable (flames). For unlabeled inputs from each domain, the merged model aligns both hidden states $\mathbf{h}_{\ell}$ (green) and logits $\mathbf{z}$ (yellow) to the corresponding expert, producing a single model that preserves all task performance.}
\label{fig:model}
\end{figure}

The key idea of Expert Merging is to learn per-layer coefficients by explicitly aligning the merged model’s internal representations and predictions to those of the experts on unlabeled inputs from their respective domains. The end-to-end training pipeline is depicted in Figure~\ref{fig:model}: the base and expert models are frozen, the coefficients are the only trainable parameters, and alignment is applied on both hidden states and logits.

\textbf{Parameterization.} Expert Merging uses one trainable coefficient per layer (panel (c) of Figure~\ref{fig:vector}). For each layer $\ell$, we set $\theta_{\text{merged}}^{\ell}=\theta_{\text{base}}^{\ell}+\sum_{k=1}^K \alpha_{k}^{\ell}\,\tau_{k}^{\ell}$, where $\alpha_{k}^{\ell}\in\mathbb{R}$ is a learnable coefficient for expert $k$ at layer $\ell$. This yields only $K\times L$ trainable coefficients, initialized with a fixed value $\bar\alpha_{k}$.

\textbf{Alignment losses.} For an input $x$, we denote the merged model’s hidden state at layer $\ell$ by $\mathbf{h}_{\ell}(x;\theta_{\text{merged}})$ and its pre-softmax logits by $\mathbf{z}(x;\theta_{\text{merged}})$. Likewise, $\mathbf{h}^{(k)}_{\ell}(x)$ and $\mathbf{z}^{(k)}(x)$ denote the corresponding values from expert $k$. Specifically, for inputs $x\in\mathcal{D}_k$ of task $k$, we minimize a combination of hidden-state and logit alignment (green and yellow dashed connections in Figure~\ref{fig:model}):
\begin{align}
\mathcal{L}^{(k)}_{\text{hid}} &\;=\; \sum_{\ell\in\mathcal{S}} \mathbb{E}_{x\sim\mathcal{D}_k} \big\| \mathbf{h}_\ell(x;\theta_{\text{merged}}) - \mathbf{h}^{(k)}_\ell(x) \big\|_2^2, \label{eq:hidden_loss}\\
\mathcal{L}^{(k)}_{\text{logit}} &\;=\; T^2\,\mathbb{E}_{x\sim\mathcal{D}_k} \operatorname{KL}\Big( \operatorname{softmax}(\mathbf{z}^{(k)}(x)/T)\;\big\|\; \operatorname{softmax}(\mathbf{z}(x;\theta_{\text{merged}})/T) \Big) \label{eq:logits_loss}, 
\end{align}
where $\mathcal{S}$ is a subset of layers used for hidden alignment (e.g., every Transformer layer’s output), and $T$ is the temperature. Hidden-state matching enforces intermediate alignment, and logit alignment promotes task-faithful outputs, which are crucial for downstream performance.

\textbf{Controllable task trade-offs.} Unlike prior trainable merging that implicitly balances tasks without offering any external control, we introduce explicit nonnegative task weights $\{\beta_k\}$ to control domain priority. Conceptually, each task contributes a weighted alignment signal: $\mathcal{L}_{\text{align}}=\sum_{k=1}^K \beta_k\big(\mathcal{L}_{\text{hid}}^{(k)} + \mathcal{L}_{\text{logit}}^{(k)}\big)$ with $\beta_k\ge 0$. Setting $\beta_k$ modulates how strongly expert $k$ is preserved, providing direct, interpretable control over cross-task trade-offs.

\textbf{Coefficient regularization.} To stabilize optimization and mitigate distribution shifts, we regularize the coefficients around the initial value $\bar\alpha_{k}$: 
\begin{equation}
\mathcal{R}(\alpha)\;=\; \frac{1}{KL} \sum_{k=1}^K \sum_{\ell=1}^L \big|\alpha_{k}^{\ell} - \bar\alpha_{k}\big|.
\end{equation}
The overall objective is $\min_{\{\alpha_{k}^{\ell}\}}\; \mathcal{L}_{\text{align}} + \gamma\,\mathcal{R}(\alpha)$, where $\gamma$ controls regularization strength. Furthermore, since the initial values have a significant impact on the optimization process, we use coefficients validated by Task Arithmetic as the initial values.

\subsection{Expert Merging++}
\label{subsec:expert-merging-plus}

Layer-wise coefficients assume uniform importance across layers and parameter groups, which is violated in practice: (i) different layer types (e.g., attention vs. MLP) have very different parameter counts; (ii) even within a type, deeper layers can be more influential; and (iii) tasks impact layers unevenly. Expert Merging++ addresses this by estimating layer importance and allocating finer-grained, chunk-wise coefficients where they matter most (panel (d) of Figure~\ref{fig:vector}).

\textbf{Layer-importance metric.} After training Expert Merging, we compute a normalized importance score per layer based on three factors: learned coefficient magnitude, task-vector weight, and parameter count. Specifically, let $s_{k}^{\ell}$ denote the task-vector weight of expert $k$ at layer $\ell$, computed from the SFT task vector (e.g., mean absolute magnitude $s_{k}^{\ell}=\operatorname{mean}(|\tau_{k}^{\ell}|)$) and normalized across layers. Then $I_\ell=\operatorname{Norm}\big( \sum_{k=1}^K |\alpha_{k}^{\ell}|\, s_{k}^{\ell}\, n_\ell \big)$, where $n_\ell=\mathrm{param\_count}(\ell)$ is the number of parameters in layer $\ell$, and $\operatorname{Norm}(\cdot)$ denotes $\ell_1$ normalization to $[0,1]$ across layers.

\textbf{Importance-guided chunk-wise coefficients.} 
As we have calculated the importance of different layers $I_\ell$, the next step is to chunk the layers according to their importance, with more chunks representing more trainable coefficients. Given a total coefficient budget $B$ for each task, we allocate to layer $\ell$ a chunk count
\begin{equation}
m_\ell\;=\; \big\lfloor B\,\frac{I_\ell^{\kappa}}{\sum_{j=1}^L I_j^{\kappa}} \big\rfloor , \qquad \kappa\ge 0, \label{eq:chunk-count}
\end{equation}
and partition its parameter tensors into $m_\ell$ parameter chunks. Here $\kappa$ is an allocation steepness hyperparameter: $\kappa=0$ yields near-uniform allocation; $\kappa=1$ allocates proportional to importance; $\kappa>1$ concentrates chunks on high-importance layers; $0<\kappa<1$ flattens the distribution. 
In implementation, we flatten the tensor and split it into $m_\ell$ contiguous chunks. Each chunk $s$ receives a distinct trainable coefficient $\alpha_{k,s}^{\ell}$ for expert $k$. 
For some low-importance layers, we set $m_\ell=0$ and use a fixed, untrainable scalar, preserving a compact parameterization.

\textbf{Objective.} We reuse the alignment objective with chunk-wise coefficients (Eq.\ref{eq:hidden_loss} and Eq.\ref{eq:logits_loss}). Let $\mathcal{S}_{\ell,s}(\cdot)$ denote a chunk-selection operator that extracts chunk $s$ from the layer $\ell$, and define $\tau_{k,s}^{\ell}=\mathcal{S}_{\ell,s}(\tau_{k}^{\ell})$. 
For each chunk $\tau_{k,s}^{\ell}$, we will have a trainable chunk-wise coefficient $\alpha_{k,s}^{\ell}$. Then
\begin{equation}
\theta_{\text{merged}}^{\ell}\;=\;\theta_{\text{base}}^{\ell}\;+\;\sum_{k=1}^K \sum_{s=1}^{m_\ell} \alpha_{k,s}^{\ell}\,\tau_{k,s}^{\ell}.
\end{equation}
For brevity set $\boldsymbol{\alpha}_{k}^{\ell}=(\alpha_{k,1}^{\ell},\ldots,\alpha_{k,m_\ell}^{\ell})$ and $\boldsymbol{\tau}_{k}^{\ell}=(\tau_{k,1}^{\ell},\ldots,\tau_{k,m_\ell}^{\ell})$, a shorthand also used in Figure~\ref{fig:model}. 
The loss is identical to Section~\ref{subsec:expert-merging}, replacing $\alpha_{k}^{\ell}$ by $\alpha_{k,s}^{\ell}$. Regularization includes a chunk-aware term $\frac{1}{BK}\sum_{k,\ell,s} \big|\alpha_{k,s}^{\ell}-\bar\alpha_{k}\big|$ which prevents the chunk-wise coefficients from deviating excessively from their initial distribution.

Importance-guided chunking allocates learnable capacity to layers that matter most, improving performance especially when there are significant differences among experts or layers (as observed in our experiments). 
Moreover, since the value of $B$ remains close to 1 in practice (0.9-1.2), even the number of chunk-wise coefficients is nearly the same as that of the layer-wise coefficients, thereby preserving strong sparsity in the merged model while improving the performance.

\section{Experiment}
\label{sec:exp}

\subsection{Setup}
\textbf{Backbones.} We evaluate our method on both LLMs and MLLMs. For LLMs, we use Mistral-7B-v0.1~\citep{chaplot2023albert} as the base model and use experts specialized for three capabilities: instruction following (Chat)\footnote{\url{https://huggingface.co/mistralai/Mistral-7B-Instruct-v0.1}}, mathematical reasoning (Math)\footnote{\url{https://huggingface.co/TIGER-Lab/MAmmoTH2-7B}}, and code generation (Code)\footnote{\url{https://huggingface.co/Nondzu/Mistral-7B-codealpaca-lora}}. For MLLMs, following prior merging methods~\citep{wei2025unifying}, we adopt InternVL2.5-1B-Instruct~\citep{chen2024expanding} and Qwen2-VL-7B-Base~\citep{wang2024qwen2} as backbones, and use experts for five capabilities: visual question answering (VQA), geometry reasoning (Geometry), chart understanding (Chart), optical character recognition (OCR), and visual grounding (Grounding). 

\textbf{Benchmarks.} For LLMs, we assess instruction following on AlpacaEval~\citep{Li_AlpacaEval_An_Automatic_2023}, mathematical reasoning on GSM8K~\citep{cobbe2021training} and MATH~\citep{hendrycks2021measuring}, and code generation on HumanEval~\citep{chen2021evaluating} and MBPP~\citep{austin2021program}. Unless otherwise noted, we report zero-shot accuracy; for code generation we use pass@1 to measure functional correctness. For MLLMs, we evaluate VQA on VizWiz~\citep{gurari2018vizwiz} and GQA~\citep{hudson2019gqa}, Geometry on MathVista~\citep{lu2024mathvista} and MATH-Vision~\citep{wang2024measuring}, Chart on ChartQA~\citep{masry2022chartqa}, OCR on TextVQA~\citep{singh2019towards} and OCRVQA~\citep{mishra2019ocr}, and Grounding on RefCOCO~\citep{kazemzadeh2014referitgame}/RefCOCO+\citep{kazemzadeh2014referitgame}/RefCOCOg~\citep{mao2016generation}. We run the MLLM evaluations using the VLMEvalKit~\citep{duan2024vlmevalkit} and LMMs-Eval~\citep{zhang2024lmms} toolchains under matched settings to ensure fair comparison. When evaluating AlpacaEval, MathVista, and MATH-Vision, we employ the GPT-4o-mini API to evaluate or extract answers from the output.

\textbf{Baselines.} We compare against strong training-free and training-based model-merging baselines: Weight Average~\citep{wortsman2022model}, Task Arithmetic (TA)~\citep{ilharco2022editing}, TIES Merging~\citep{yadav2023ties}, DARE~\citep{yu2024language}, TSV Merging~\citep{gargiulo2025task}, Iso-C~\citep{marczak2025no}, WUDI Merging~\citep{cheng2025whoever}, WUDI v2~\citep{wei2025unifying}, and AdaMerging/AdaMerging++~\citep{AdaMerging_ICLR_2024}. Detailed baseline descriptions are provided in Appendix~\ref{app:exp-setting}.

\textbf{Implementation details.} 
For each task, we sample 5–10 unlabeled examples from the training or test sets as the dataset. We first perform Expert Merging, and then apply Expert Merging++ using the learned coefficients. All experiments are repeated five times and results are averaged, using $8\times$ NVIDIA V100 GPUs. Additional hyperparameters and details are provided in Appendix~\ref{app:impl}.

\subsection{Main Results}
Across both the MLLM and LLM backbones, our approach delivers the strongest overall performance while using only unlabeled data for calibration. On InternVL (Table~\ref{tab:main-internvl}), Expert Merging++ attains the best average of 58.45, outperforming the best prior merging method (+1.49 over WUDI v2) and even exceeding supervised Mixture Training (+0.79). The gains are particularly pronounced on grounding benchmarks, where \emph{Expert Merging/++} achieves 80.05/80.53 on RefCOCO, 73.85/74.37 on RefCOCO+, and 79.04/79.31 on RefCOCOg, substantially surpassing all baselines. At the same time, VQA performance remains competitive (e.g., 31.16 on VizWiz, the best in table) and geometry scores are strong (26.32 on MATH-Vision with Expert Merging), indicating that the method effectively balances recognition, reasoning, and localization. While certain OCR-centric metrics (e.g., OCRVQA) favor Mixture Training, the overall average and grounding improvements demonstrate superior cross-domain integration. Additional analysis and results are provided in Appendix~\ref{app:exp-results}.

\begin{table}[t]
\centering
\caption{
Main results on InternVL2.5 across multiple tasks.}
\label{tab:main-internvl}
\begingroup
\setlength{\tabcolsep}{2.5pt}
\renewcommand{\arraystretch}{1.05}
\footnotesize
\resizebox{\linewidth}{!}{%
\begin{tabular}{l cc @ {\hspace{6pt}} cc @ {\hspace{6pt}} c @ {\hspace{6pt}} cc @ {\hspace{6pt}} ccc @ {\hspace{6pt}} c}
\toprule
 & \multicolumn{2}{c}{\textbf{VQA}} & \multicolumn{2}{c}{\textbf{Geometry}} & \multicolumn{1}{c}{\textbf{Chart}} & \multicolumn{2}{c}{\textbf{OCR}} & \multicolumn{3}{c}{\textbf{Grounding}} & \\
\cmidrule(lr){2-3} \cmidrule(lr){4-5} \cmidrule(lr){6-6} \cmidrule(lr){7-8} \cmidrule(lr){9-11}
\textbf{Method} & \textbf{VizWiz} & \textbf{GQA} & \textbf{MathVista} & \textbf{MATH-Vision} & \textbf{ChartQA} & \textbf{TextVQA} & \textbf{OCRVQA} & \textbf{RefCOCO} & \textbf{RefCOCO+} & \textbf{RefCOCOg} & \textbf{Avg.} \\
\midrule

\textbf{InternVL2.5-Instruct} & 29.15 & 54.62 & 46.80 & 18.42 & 69.48 & 72.51 & 41.08 & 71.69 & 65.41 & 67.40 & 53.65 \\ 
\midrule
\textbf{VQA Expert} & 30.58 & \underline{60.91} & 35.50 & 17.11 & 48.76 & 63.68 & 36.04 & - & - & - & 41.79 \\ 
\textbf{Geometry Expert} & 13.45 & 32.80 & \textbf{55.20} & 25.00 & 51.76 & 56.91 & 35.35 & 24.73 & 19.61 & 23.84 & 33.86 \\ 
\textbf{Chart Expert} & 20.16 & 40.39 & 23.84 & 10.53 & 69.52 & 54.36 & 34.83 & - & - & - & 36.23 \\ 
\textbf{OCR Expert} & 12.40 & 22.22 & 23.31 & 10.53 & 36.88 & 73.00 & 54.79 & 73.65 & 68.01 & 69.10 & 44.38 \\ 
\textbf{Grounding Expert} & 19.09 & 25.88 & 28.91 & 14.47 & 41.32 & 58.39 & \textbf{74.87} & 76.67 & 71.35 & 70.09 & 48.10 \\ 
\midrule
\textbf{Weight Average} & 29.96 & 54.89 & 49.60 & 18.42 & 71.64 & 74.54 & 41.86 & 52.62 & 45.29 & 52.39 & 49.12 \\ 
\textbf{Task Arithmetic} & 30.67 & 56.34 & 45.36 & 21.05 & \underline{72.88} & 76.26 & 43.39 & 74.90 & 68.15 & 72.75 & 56.17 \\ 
\textbf{TIES Merging} & 30.63 & 56.48 & 44.50 & 23.68 & 72.28 & \underline{76.29} & 44.01 & 76.01 & 68.45 & 73.65 & 56.59 \\ 
\textbf{TA w/ DARE} & 30.61 & 56.48 & 48.45 & 21.05 & \textbf{73.08} & \textbf{76.30} & 43.03 & 74.94 & 68.07 & 73.02 & 56.50 \\ 
\textbf{TIES w/ DARE} & 30.65 & 56.11 & 43.85 & \underline{27.63} & 72.72 & 76.19 & 43.33 & 75.10 & 68.48 & 73.55 & 56.76 \\ 
\textbf{TSV Merging} & \underline{31.15} & 56.67 & 52.45 & \textbf{28.95} & 70.56 & 75.66 & 45.38 & 65.19 & 58.51 & 59.17 & 54.36 \\ 
\textbf{Iso-C} & 28.21 & 55.36 & 48.96 & 21.05 & 70.56 & 69.34 & 46.51 & 72.72 & 66.56 & 68.50 & 54.77 \\ 

\textbf{WUDI Merging} & 30.22 & 56.07 & 52.85 & 19.74 & 70.00 & 75.48 & 45.74 & 75.61 & 69.40 & 73.53 & 56.86 \\

\textbf{WUDI v2} & 31.00 & 57.23 & 52.61 & 21.05 & 65.32 & 76.20 & 45.83 & 76.30 & 69.91 & 74.16 & 56.96 \\

\textbf{AdaMerging} & 31.06 & 57.20 & \underline{55.10} & 25.00 & 62.32 & 76.20 & 44.76 & 75.30 & 68.42 & 73.22 & 56.85 \\ 

\textbf{AdaMerging++} & 30.94 & 57.18 & 51.77 & 19.74 & 66.20 & 76.02 & 45.93 & 76.45 & 70.47 & 74.41 & 56.91 \\ 

\textbf{Mixture Training} & 29.79 & \textbf{61.33} & 52.83 & 23.68 & 70.32 & 72.96 & \underline{60.25} & 72.06 & 65.93 & 67.46 & 57.66 \\ 
\midrule
\textbf{Expert Merging} & \textbf{31.16} & 56.77 & 47.71 & 26.32 & 65.28 & 75.70 & 45.31 & \underline{80.05} & \underline{73.85} & \underline{79.04} & \underline{58.11} \\ 
\textbf{Expert Merging++} & 31.14 & 56.53 & 52.35 & 22.37 & 65.92 & 76.06 & 46.00 & \textbf{80.53} & \textbf{74.37} & \textbf{79.31} & \textbf{58.45} \\

\bottomrule
\end{tabular}%
}
\endgroup
\end{table}

Results on Qwen2-VL (Table~\ref{tab:main-qwen2vl}) show the same pattern. Expert Merging++ achieves the best average of 63.63, outperforming the strongest prior merging baseline (+1.00 over WUDI v2) and Mixture Training (+1.40). It sets or matches state-of-the-art results on several domains, including MATH-Vision (44.74) and TextVQA (81.65), while delivering excellent grounding (e.g., 79.00 on RefCOCOg). Although some methods peak on individual tasks (e.g., TA+DARE on ChartQA), they do so at the expense of other domains, whereas \emph{Expert Merging/++} maintains consistently high performance across VQA, geometry, OCR, and grounding, yielding the best overall trade-off.

\begin{table}[t]
\centering
\caption{
Main results on Qwen2-VL across multiple tasks.}
\label{tab:main-qwen2vl}
\begingroup
\setlength{\tabcolsep}{2.5pt}
\renewcommand{\arraystretch}{1.05}
\footnotesize
\resizebox{\linewidth}{!}{%
\begin{tabular}{l cc @ {\hspace{6pt}} cc @ {\hspace{6pt}} c @ {\hspace{6pt}} cc @ {\hspace{6pt}} ccc @ {\hspace{6pt}} c}
\toprule
 & \multicolumn{2}{c}{\textbf{VQA}} & \multicolumn{2}{c}{\textbf{Geometry}} & \multicolumn{1}{c}{\textbf{Chart}} & \multicolumn{2}{c}{\textbf{OCR}} & \multicolumn{3}{c}{\textbf{Grounding}} & \\
\cmidrule(lr){2-3} \cmidrule(lr){4-5} \cmidrule(lr){6-6} \cmidrule(lr){7-8} \cmidrule(lr){9-11}
\textbf{Method} & \textbf{VizWiz} & \textbf{GQA} & \textbf{MathVista} & \textbf{MATH-Vision} & \textbf{ChartQA} & \textbf{TextVQA} & \textbf{OCRVQA} & \textbf{RefCOCO} & \textbf{RefCOCO+} & \textbf{RefCOCOg} & \textbf{Avg.} \\
\midrule

\textbf{Qwen2-VL-Base} & 5.52 & 5.39 & 47.85 & 23.68 & 0.36 & 20.22 & 1.07 & 45.32 & 37.55 & 31.26 & 21.82 \\ 
\midrule
\textbf{VQA Expert} & 41.38 & \textbf{62.60} & 33.71 & 28.94 & 66.56 & 80.21 & 55.33 & 39.31 & 32.71 & 38.01 & 47.88 \\ 
\textbf{Geometry Expert} & 35.57 & 44.63 & 42.50 & 28.95 & 14.56 & 73.95 & 45.96 & 5.57 & 2.31 & 3.90 & 29.79 \\ 
\textbf{Chart Expert} & 38.58 & 24.24 & 49.28 & 32.89 & 61.08 & 79.75 & 63.67 & 46.28 & 36.67 & 34.06 & 46.65 \\ 
\textbf{OCR Expert} & 28.38 & 37.53 & 31.81 & 13.16 & 57.40 & 70.50 & 64.68 & 0.59 & 0.46 & 0.26 & 30.48 \\ 
\textbf{Grounding Expert} & 38.60 & 32.92 & 36.17 & 19.74 & 18.08 & 75.05 & 48.27 & 72.14 & 65.33 & 66.48 & 47.28 \\ 
\midrule
\textbf{Weight Average} & 41.47 & 57.33 & \underline{50.21} & 34.21 & 59.56 & 81.09 & 57.85 & 80.72 & 65.37 & 77.68 & 60.55 \\ 
\textbf{Task Arithmetic} & 40.52 & 62.31 & 40.36 & 26.31 & \underline{79.67} & 81.09 & 59.50 & 75.96 & 61.33 & 75.85 & 60.29 \\ 
\textbf{TIES Merging} & 41.38 & 59.08 & 46.87 & 34.21 & 67.24 & 81.42 & 58.53 & 80.63 & 65.36 & 77.65 & 61.24 \\ 
\textbf{TA w/ DARE} & 40.64 & \underline{62.38} & 40.67 & 26.31 & \textbf{79.76} & 81.04 & 59.34 & 75.83 & 61.41 & 75.80 & 60.32 \\ 
\textbf{TIES w/ DARE} & 41.63 & 59.96 & 45.72 & 35.53 & 70.68 & 81.53 & 59.63 & \underline{80.73} & 65.65 & 77.77 & 61.88 \\ 
\textbf{TSV Merging} & 41.43 & 57.31 & \textbf{51.05} & 34.21 & 59.44 & 81.25 & 57.81 & 80.71 & 65.34 & 77.76 & 60.63 \\ 
\textbf{Iso-C} & 12.31 & 13.44 & 39.96 & 27.63 & 2.80 & 30.05 & 6.12 & 53.68 & 38.96 & 41.90 & 26.69 \\ 
\textbf{WUDI Merging} & 37.19 & 56.45 & 42.96 & 27.63 & 67.84 & 79.92 & \textbf{65.56} & 76.25 & 60.72 & 71.99 & 58.65 \\ 

\textbf{WUDI v2} & 41.13 & 61.17 & 46.98 & 38.16 & 74.64 & 79.54 & 60.03 & 80.43 & 65.85 & 78.42 & 62.63 \\

\textbf{AdaMerging} & \underline{41.83} & 60.28 & 49.08 & 35.53 & 71.68 & 81.52 & 60.22 & 53.31 & 47.27 & 57.62 & 55.83 \\ 
\textbf{AdaMerging++} & 41.67 & 60.09 & 47.82 & 35.53 & 69.76 & \underline{81.59} & 60.42 & 80.57 & 65.98 & 78.74 & 62.22 \\ 

\textbf{Mixture Training} & \textbf{44.09} & 62.18 & 46.02 & 19.73 & 70.04 & 78.38 & \underline{65.42} & \textbf{82.89} & \textbf{77.87} & 75.63 & 62.23 \\ 

\midrule

\textbf{Expert Merging} & 41.54 & 60.51 & 48.80 & \underline{42.11} & 73.00 & 81.41 & 60.19 & 80.64 & 66.15 & \textbf{79.02} & \underline{63.34} \\ 
\textbf{Expert Merging++} & 41.49 & 60.42 & 49.33 & \textbf{44.74} & 72.60 & \textbf{81.65} & 60.19 & 80.71 & \underline{66.21} & \underline{79.00} & \textbf{63.63} \\

\bottomrule
\end{tabular}%
}
\endgroup
\end{table}

\begin{wraptable}[14]{r}{0.42\textwidth}
\centering
\vspace{-10pt} 
\caption{Main results on Mistral across multiple tasks.}
\label{tab:main-mistral}
\begingroup
\setlength{\tabcolsep}{1.6pt}
\renewcommand{\arraystretch}{1.05}
\scriptsize
\resizebox{\linewidth}{!}{%
\begin{tabular}{l c @ {\hspace{6pt}} cc @ {\hspace{6pt}} cc @ {\hspace{6pt}} c}
\toprule
 & \multicolumn{1}{c}{\textbf{Chat}} & \multicolumn{2}{c}{\textbf{MATH}} & \multicolumn{2}{c}{\textbf{Code}} & \\
\cmidrule(lr){2-2} \cmidrule(lr){3-4} \cmidrule(lr){5-6}
\textbf{Method} & \textbf{Alpaca Eval} & \textbf{GSM8k} & \textbf{Math} & \textbf{HumanEval} & \textbf{MBPP} & \textbf{Avg.} \\
\midrule

\textbf{Mistral-7B-v0.1} & 4.87 & 13.57 & 5.10 & 30.49 & 40.80 & 18.97 \\ 
\midrule
\textbf{Chat Expert} & 71.37 & 41.09 & 7.26 & 35.37 & 32.20 & 37.46 \\ 
\textbf{Math Expert} & 53.18 & \textbf{63.38} & \textbf{25.22} & 20.73 & 32.40 & 38.98 \\ 
\textbf{Code Expert} & 69.90 & 30.78 & 4.84 & \textbf{54.88} & 38.20 & 39.72 \\ 
\midrule
\textbf{Task Arithmetic} & 74.18 & 52.69 & 13.16 & 45.12 & 26.00 & 42.23 \\ 
\textbf{TIES Merging} & 71.66 & 53.68 & 13.86 & 38.41 & 40.00 & 43.52 \\ 
\textbf{TA w/ DARE} & 74.92 & 56.71 & 13.22 & 43.29 & \textbf{43.40} & 42.71 \\ 
\textbf{TIES w/ DARE} & \underline{76.33} & 52.08 & 12.00 & 34.76 & \textbf{43.40} & 43.71 \\ 
\textbf{TSV Merging} & 60.61 & 47.38 & 11.18 & 37.20 & \underline{42.20} & 39.71 \\ 
\textbf{Iso-C} & 69.76 & 52.39 & 12.82 & 32.32 & 40.60 & 41.58 \\ 
\textbf{WUDI Merging} & 64.93 & 53.37 & 14.04 & 32.93 & \textbf{43.40} & 41.73 \\ 
\textbf{WUDI v2} & 73.84 & 59.06 & \underline{14.88} & 39.02 & 40.60 & 45.48 \\ 
\textbf{AdaMerging} & 75.94 & 54.81 & 12.24 & 43.29 & 39.00 & 45.06 \\ 
\textbf{AdaMerging++} & 75.94 & 54.81 & 12.24 & 43.29 & 39.20 & 45.10 \\ 
\midrule
\textbf{Expert Merging} & \textbf{77.84} & 58.68 & 13.88 & 46.34 & 40.40 & \underline{47.43} \\ 
\textbf{Expert Merging++} & 75.51 & \underline{59.44} & 13.94 & \underline{53.66} & 41.00 & \textbf{48.71} \\ 

\bottomrule
\end{tabular}%
}
\relax
\endgroup

\end{wraptable}

For Mistral-7B (Table~\ref{tab:main-mistral}), Expert Merging++ attains the highest average (48.71), improving over strong baselines such as WUDI v2 (+3.23) and AdaMerging++ (+3.61). It achieves the best Code score among merged models (53.66 on HumanEval), with strong performance on AlpacaEval (75.51) and GSM8K (59.44). Notably, Expert Merging attains the highest Chat score overall (77.84 on AlpacaEval). Our merged models approach or match specialist experts within their domains while consolidating all capabilities into a single model, validating the effectiveness of our coefficient learning and alignment objectives.

Together, these results support our key claim: aligning both hidden states and logits with per-layer coefficients yields a robust, training-efficient path to unify domain experts. The chunk-wise coefficients of Expert Merging++ (guided by the importance analysis in Section~\ref{subsec:importance}) further improve cross-domain balance with negligible overhead, leading to state-of-the-art averages across backbones without supervised finetuning.

\subsection{Layer-Wise Importance Analysis}
\label{subsec:importance}
Merging performance is not uniform across layers or depths. To investigate this heterogeneity, we examined the coefficient magnitudes and parameter sizes of each layer across different component types (attention projections Q/K/V/O and MLP Gate/Up/Down) and depths (early, middle, late). We then combined these statistics with the task vector weights to derive a per-layer importance score, quantifying each layer’s contribution to the final merged model.

\begin{figure}[t]
\centering
\includegraphics[width=0.8\linewidth]{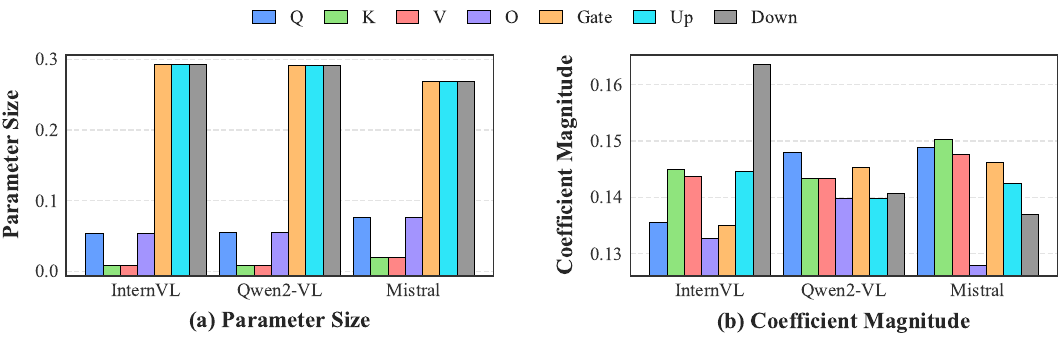}
\caption{Model-wise parameter size and learned coefficient across backbones after normalizing. Full stage-wise trends for panel (b) are provided in Appendix (Figure~\ref{fig:coefficients_share_by_stage}).}
\label{fig:param_size_vs_coefficients_share_modelwise}
\end{figure}

\textbf{Parameter size across backbones.} Despite architectural differences, the distribution of parameters by submodule type is highly consistent (panel (a) of Figure~\ref{fig:param_size_vs_coefficients_share_modelwise}). In Qwen and InternVL, the MLP dominates capacity: Gate/Up/Down together account for roughly 87\% of the parameters, while attention uses the remaining 13\%. Mistral has a slightly larger attention parameters (about 19\%) but the same qualitative pattern. This structure suggests that most of the representational capacity resides in the MLP, with Q/O providing relatively smaller but non-negligible attention-side capacity.

\textbf{Learned coefficient magnitude.} After normalizing the learned coefficients within each model, we aggregate them across submodule to obtain panel (b) of Figure~\ref{fig:param_size_vs_coefficients_share_modelwise}. The magnitudes are fairly balanced overall, but the leading submodule varies by backbone: Down is largest for InternVL, Q for Qwen2-VL, and K for Mistral. The O projection is consistently the smallest across all three. Notably, Down exhibits the greatest variance; it is highest on InternVL while being lowest on Mistral. This indicates backbone-specific allocation rather than a universal ranking. This pattern contrasts with panel (a), where MLP (Gate/Up/Down) clearly dominates by parameter size.

\begin{figure}[t]
\centering
\includegraphics[width=\linewidth]{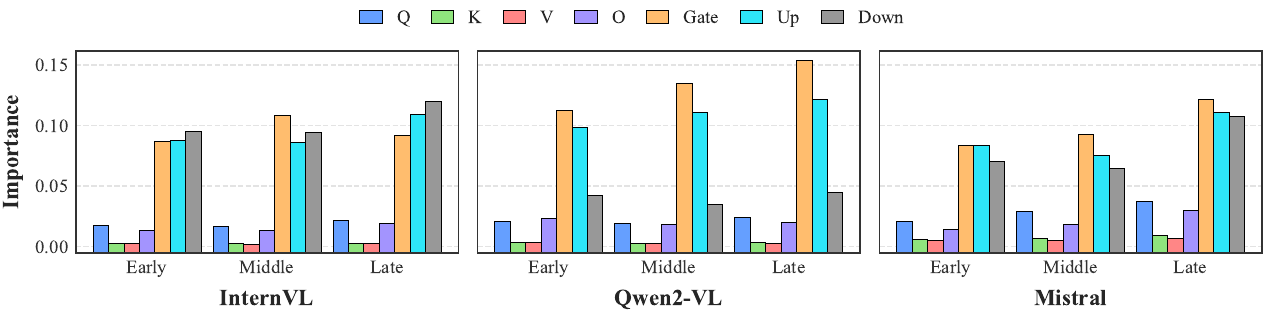}
\caption{Layer importance by stage and submodule across backbones.}
\label{fig:importance_by_stage}
\end{figure}

\textbf{Importance: where capacity actually matters.} 
As described in Section~\ref{subsec:expert-merging-plus}, we compute layer importance by stage and submodule across backbones, with results shown in Figure~\ref{fig:importance_by_stage}. First, MLP components dominate: Gate and Up consistently rank as the two largest contributors, followed by Down; attention K and V contribute the least; Q and O fall in between. Second, importance concentrates in the late-stage third of the network for all types, with monotonic or near-monotonic increases from early to late blocks (particularly pronounced for Mistral). Third, backbone-specific nuances remain, InternVL’s Down-projection is the single largest contributor, whereas Mistral distributes MLP importance more evenly between Gate and Up, but the overarching ranking is consistent.

\textbf{Takeaways for Expert Merging++.} The analysis reveals substantial variation across submodules and depths in different architectures, indicating that the prior practice of assigning a single coefficient per layer overlooks critical inter-layer heterogeneity. This observation motivates Expert Merging++, which replaces the layer-wise coefficient with chunk-wise coefficients, enabling finer-grained control in high-importance layers. In particular, high-importance layers exhibit larger merged task vectors and influence more parameters, explaining why chunking only these layers recovers nearly all the benefits of dense chunking, whereas chunking low-importance layers yields little gain.

\subsection{Ablation Studies}

We study how each component of Expert Merging and Expert Merging++ contributes to performance across the two MLLM backbones and the Mistral LLM. For brevity, we report a consolidated summary in Table~\ref{tab:ablation}, and defer the full per-benchmark ablation tables to Appendix~\ref{app:full-ablation}.

\begin{wraptable}[8]{r}{0.42\textwidth}
\centering
\vspace{-22pt}

\caption{Ablations on key design choices.}
\label{tab:ablation}
\begingroup
\small
\resizebox{\linewidth}{!}{%
\begin{tabular}{lccc}
\toprule
\textbf{Variant} & \textbf{InternVL} & \textbf{Qwen2-VL} & \textbf{Mistral} \\
\midrule
Expert Merging & \underline{58.11} & \underline{63.34} & \underline{47.43} \\
\quad w/o hidden alignment & 57.75 & 62.43 & 47.10 \\
\quad w/o logit alignment & 54.72 & 62.71 & 43.62 \\
\quad w/o coefficient regularization & 47.57 & 62.17 & 40.81 \\
\quad w/o task trade-offs & 57.71 & 62.69 & 46.80 \\
\midrule
Expert Merging++ & \textbf{58.45} & \textbf{63.63} & \textbf{48.71} \\
\quad w/o chunking & \underline{58.11} & \underline{63.34} & \underline{47.43} \\
\quad chunk all layers (2x) & 56.76 & 62.61 & 46.82 \\
\bottomrule
\end{tabular}%
}
\endgroup
\end{wraptable}

\textbf{Alignment objectives.} Removing either alignment term hurts performance across all backbones, confirming that matching both internal representations and predictive distributions is beneficial. On InternVL and Mistral, logit alignment accounts for the larger gains (\mbox{$-3.39$} and \mbox{$-3.81$} average without it), while hidden-state alignment provides additional improvements. On Qwen2-VL, the two terms contribute comparably. Qualitatively, hidden alignment stabilizes cross-domain behavior (notably OCR and grounding on InternVL), while logit knowledge aligning brings task-faithful outputs.

\textbf{Coefficient regularization.} The coefficient regularizer is critical for stability. Without it, averages drop markedly (\mbox{$-10.54$} on InternVL, \mbox{$-6.62$} on Mistral, \mbox{$-1.17$} on Qwen2-VL), accompanied by pronounced degradation on grounding metrics (e.g., RefCOCO/+/g on InternVL). This supports our design in Section~\ref{sec:method}: regularization curbs coefficient drift during unsupervised alignment and prevents catastrophic forgetting in high-capacity, late layers.

\textbf{Controllable task trade-offs.} Removing explicit task weights consistently reduces averages (\mbox{$-0.4$ to $-0.7$}). Beyond the modest aggregate gains, the weights enable practical control over cross-domain balance, aligning with our goal of exposing interpretable knobs for deployment.

\textbf{Chunk-wise coefficients.} Expert Merging++ extends layer-wise coefficients (Expert Merging) by assigning chunk-wise coefficients, yielding further improvements: +0.34 on InternVL, +0.29 on Qwen2-VL, and +1.28 on Mistral. These gains concentrate in regions where our importance analysis (Section~\ref{subsec:importance}) indicates capacity matters most, namely late-stage MLP blocks. For example, chunking substantially boosts code generation on Mistral (HumanEval \mbox{+7.3} points) while preserving or slightly improving other domains. In contrast, indiscriminately chunking all layers (2× coefficients) consistently degrades performance across backbones. This underscores that targeted capacity allocation, rather than uniform over-parameterization, is crucial for addressing inter-layer heterogeneity, and further validates both our importance metric and the chunk-wise design of Expert Merging++.

\textbf{Summary.} The ablations validate our core choices: (1) alignment losses both help (logits stronger on InternVL/Mistral, hidden states aid robustness); (2) coefficient regularization is critical for stability; (3) task weights enable controllable trade-offs; and (4) chunk-wise coefficients in Expert Merging++ deliver further gains. Together, these components yield the best balance between accuracy and stability and deliver consistent cross-domain improvements.

\section{Conclusion and Future Work}
\label{sec:conclusion}

We introduce Expert Merging, a label-free, training-light approach that learns layer-wise coefficients to align hidden states and logits, and Expert Merging++ which adds importance-guided, chunk-wise capacity. Across InternVL, Qwen2-VL, and Mistral, they consistently surpass training-free and prior training-based baselines; Expert Merging++ provides further gains and in some cases exceeds supervised Mixture Training, with ablations confirming each design choice.
In future work, we plan to further explore model merging solutions for heterogeneous models.

\bibliography{iclr2026_conference}
\bibliographystyle{iclr2026_conference}

\newpage

\appendix

\section{Use of Large Language Models}
\label{app:llm-usage}
We used large language models (LLMs) to aid and polish the writing of this manuscript. Concretely, we used LLMs to suggest improvements to grammar, phrasing, and clarity for select passages. 

\section{Experiment Setting}
\label{app:exp-setting}

\subsection{Baselines}
We compare against the following baselines:
\begin{itemize}
  \item \textbf{Weight Averaging}~\citep{wortsman2022model}: simply averages the weights (or deltas) of models fine-tuned on different tasks.
  \item \textbf{Task Arithmetic (TA)}~\citep{ilharco2022editing}: define task vectors $\tau_{k}=\theta_{k}-\theta_{\text{base}}$, and the merged model follows the notation in Section~\ref{subsec:prelim} as $\theta_{\text{TA}}=\theta_{\text{base}}+\sum_{k=1}^{K} \lambda_{k}\,\tau_{k}$ (with predefined $\{\lambda_k\}$). \textbf{DARE}~\citep{yu2024language} randomly drops redundant task vectors and rescales the remaining ones to mitigate interference.
  \item \textbf{TIES Merging}~\citep{yadav2023ties}: a sparsification-based pipeline with Trim, Elect Sign, and Disjoint Merge to produce a merged vector $\tau_{m}$; the final model is $\theta_{\text{base}}+\lambda\,\tau_{m}$ with $\lambda$ tuned; we also evaluate \textbf{TIES+DARE}.
  \item \textbf{TSV Merging}~\citep{gargiulo2025task}: an SVD-based method that measures singular task interference and reduces it via decorrelation formulated as an orthogonal Procrustes problem.
  \item \textbf{Iso-C}~\citep{marczak2025no}: isotropic merging that flattens the singular value spectrum of task matrices and improves alignment between singular components.
  \item \textbf{WUDI Merging}~\citep{cheng2025whoever}: optimization-based merging that treats task vectors as spanning an approximate linear subspace and minimizes layer-wise interference between the merged vector $\tau_{m,\ell}$ and each task vector $\tau_{i,\ell}$ using only task vectors (no labels).
  \item \textbf{WUDI v2}~\citep{wei2025unifying}: adds an SVD-based component to decompose task matrices, yielding stronger multi-domain performance.
  \item \textbf{AdaMerging}~\citep{AdaMerging_ICLR_2024}: training-based approach that learns small, typically per-layer coefficients on unlabeled calibration data to combine experts.
  \item \textbf{AdaMerging++}~\citep{AdaMerging_ICLR_2024}: a TIES-based variant that learns per-layer coefficients for each task vector \emph{within the TIES-Merging pipeline}, improving robustness to conflicts.
  \item \textbf{Mixture Training}: supervised multi-task finetuning on the union of data, serving as an upper bound reference when available.
\end{itemize}

\paragraph{Analysis.}
Across backbones (Tables~\ref{tab:main-internvl}, \ref{tab:main-qwen2vl}, and \ref{tab:main-mistral}), training-free methods exhibit clear trade-offs. TA/DARE variants often peak on a few domains (e.g., ChartQA) but underperform on grounding or geometry tasks, while sparsification- and SVD-based methods (TIES/TSV/Iso-C) reduce conflicts yet still lag on cross-domain balance. Optimization-oriented WUDI/WUDI~v2 deliver stronger macro averages among training-free baselines but remain below \emph{Expert Merging/++}, especially on grounding (MLLMs) and code (LLMs). Notably, Mixture Training is not a strict upper bound in our setting: \emph{Expert Merging++} surpasses it on macro average for both InternVL and Qwen2-VL, echoing the main text that careful, unlabeled calibration can rival or exceed supervised multi-task fine-tuning.

\subsection{Implementation details}
\label{app:impl}
\textbf{Backbones and experts.} We consider both LLMs and MLLMs. For LLMs, the backbone is Mistral-7B-v0.1 with three domain experts: instruction following (Chat), mathematical reasoning (Math), and code generation (Code). For MLLMs, following prior multimodal merging work~\citep{wei2025unifying}, we use InternVL2.5-1B-Instruct and Qwen2-VL-7B-Base as backbones, with five capability experts: VQA, Geometry, Chart, OCR, and Grounding. Expert checkpoints follow the sources referenced in Section~\ref{sec:exp}.

\paragraph{Hyperparameters.}
We report the key hyperparameters used; other settings follow defaults and are omitted due to space.

\begin{itemize}
  \item \textbf{Train samples per task (\boldmath$N$).} InternVL2.5: \textbf{5}; Mistral-7B-v0.1: \textbf{5}; Qwen2-VL-7B-Base: \textbf{10}.
  \item \textbf{Initialization (\boldmath$\bar\alpha$ prior).} InternVL2.5: \textbf{0.1}; Mistral-7B-v0.1: \textbf{0.3}; Qwen2-VL-7B-Base: \textbf{0.3}.
  \item \textbf{Hidden alignment layers (\boldmath$\mathcal{S}$).} For all models, we supervise only the \emph{middle} Transformer layer’s hidden state, i.e., $|\mathcal{S}|{=}1$ with layer index $\lfloor L/2 \rfloor$.
  \item \textbf{Coefficient regularization weight (\boldmath$\gamma$).} InternVL2.5: \textbf{5}; Mistral-7B-v0.1: \textbf{0.8}; Qwen2-VL-7B-Base: \textbf{0.8}.
  \item \textbf{Total coefficient budget (\boldmath$B$).} Set to \textbf{$0.9$--$1.2\times L$} to reduce trainable parameters and encourage sparsity.
  \item \textbf{Allocation steepness (\boldmath$\kappa$).} Set to \textbf{1.2} for each model to accentuate differences in layer importance.
\end{itemize}

\section{Experiment Results}
\label{app:exp-results}

\subsection{Layer-Wise Importance Analysis}
\label{app:coefficients-by-stage}

This section provides the full stage-wise (early/middle/late) breakdown of learned coefficients for each submodule and backbone, complementing the model-wise aggregation in Figure~\ref{fig:param_size_vs_coefficients_share_modelwise}b.

\begin{figure}[h]
\centering
\includegraphics[width=\linewidth]{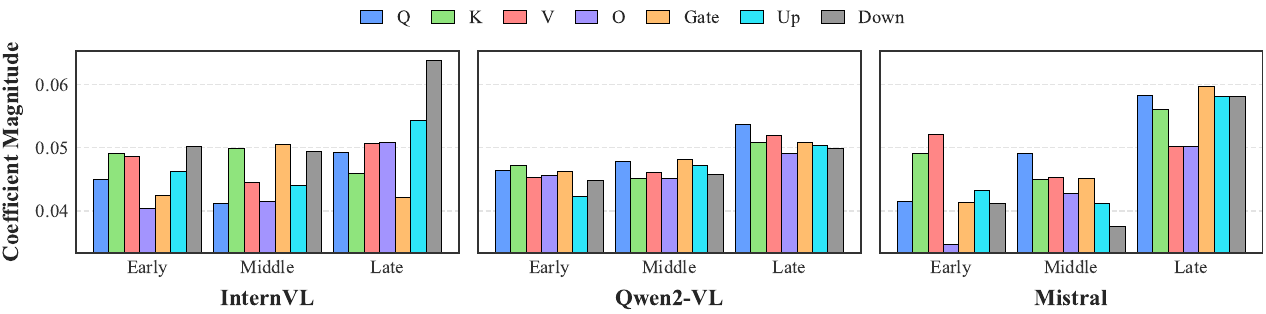}
\caption{Learned coefficient share across depth (early/middle/late) for each submodule. Late-stage blocks consistently receive larger shares for both attention and MLP; InternVL shows a stronger late-stage skew on the MLP down-projection, while Qwen’s shift is milder.}
\label{fig:coefficients_share_by_stage}
\end{figure}

\textbf{Learned coefficients across depth.} Coefficients learned by \emph{Expert Merging} are distinctly non-uniform over depth as Figure~\ref{fig:coefficients_share_by_stage} shown. Across all three backbones, late-stage blocks receive larger shares than early ones for both attention and MLP. For example, on Mistral the Q and Gate shares grow from about $\sim$0.04 (early) to $\sim$0.06 (late). InternVL shows a pronounced late-stage skew on the MLP down-projection, while Qwen exhibits a milder early-to-late increase. Averaged over types, model-wise coefficient shares remain relatively even (roughly 0.13–0.16 per type), suggesting that raw coefficient fractions alone weakly indicate capacity hot spots unless viewed alongside parameterization across depth.

\textbf{Interpretation.} The late-layer emphasis aligns with the role of deeper blocks in consolidating task-specific signals (e.g., cross-modal grounding in MLLMs and semantic control in LLMs). This observation motivated the chunk-wise refinement in \emph{Expert Merging++}, which reallocates more capacity primarily within later regions. Consistently, Appendix~\ref{app:full-ablation} shows targeted chunking improves grounding/OCR on InternVL and boosts code/math on Mistral with minimal impact on VQA.

\subsection{Ablation Studies}
\label{app:full-ablation}

This section provides the complete ablation results for each backbone, expanding the summarized findings reported in Table~\ref{tab:ablation}. See Tables~\ref{tab:ablation-internvl}, \ref{tab:ablation-qwen2vl}, and \ref{tab:ablation-mistral} for detailed results.

\paragraph{Summary.}
Three consistent trends emerge across backbones: (i) \emph{logit alignment is essential}—removing it causes the largest average drops and notably hurts geometry/grounding on MLLMs and code on LLMs; (ii) \emph{coefficient regularization is critical for stability}—without it, performance collapses on InternVL and Mistral, indicating over-allocation and interference; and (iii) \emph{chunking helps when guided by importance}—the targeted chunk-wise variant in \emph{Expert Merging++} yields macro-average gains, whereas naively chunking all layers degrades multiple domains.

\begin{table}[t]
\centering
\caption{Ablation studies on InternVL2.5. We measure the effect of each component in Expert Merging and the chunking strategy in Expert Merging++. Missing entries correspond to experiments still in progress.}
\label{tab:ablation-internvl}
\begingroup
\setlength{\tabcolsep}{2.5pt}
\renewcommand{\arraystretch}{1.05}
\footnotesize
\resizebox{\linewidth}{!}{%
\begin{tabular}{l cc @ {\hspace{6pt}} cc @ {\hspace{6pt}} c @ {\hspace{6pt}} cc @ {\hspace{6pt}} ccc @ {\hspace{6pt}} c}
\toprule
 & \multicolumn{2}{c}{\textbf{VQA}} & \multicolumn{2}{c}{\textbf{Geometry}} & \multicolumn{1}{c}{\textbf{Chart}} & \multicolumn{2}{c}{\textbf{OCR}} & \multicolumn{3}{c}{\textbf{Grounding}} & \\
\cmidrule(lr){2-3} \cmidrule(lr){4-5} \cmidrule(lr){6-6} \cmidrule(lr){7-8} \cmidrule(lr){9-11}
\textbf{Variant} & \textbf{VizWiz} & \textbf{GQA} & \textbf{MathVista} & \textbf{MATH-Vision} & \textbf{ChartQA} & \textbf{TextVQA} & \textbf{OCRVQA} & \textbf{RefCOCO} & \textbf{RefCOCO+} & \textbf{RefCOCOg} & \textbf{Avg.} \\
\midrule
\textbf{Expert Merging} & \underline{31.16} & 56.77 & 47.71 & \underline{26.32} & 65.28 & 75.70 & \underline{45.31} & \underline{80.05} & \underline{73.85} & \underline{79.04} & \underline{58.11} \\
\quad w/o hidden alignment & 31.07 & 56.81 & 49.90 & \textbf{27.63} & 64.52 & 75.55 & 45.15 & 78.06 & 71.92 & 76.98 & 57.75  \\
\quad w/o logit alignment & \textbf{31.53} & \textbf{57.08} & 48.82 & 22.37 & \textbf{67.16} & \underline{75.82} & 44.76 & 69.81 & 63.05 & 66.83 & 54.72 \\
\quad w/o task trade-offs & \underline{31.16} & 56.72 & 51.17 & \underline{26.32} & 64.60 & 75.52 & 45.05 & 78.12 & 71.65 & 76.82 & 57.71 \\
\quad w/o coefficient regularization & 29.63 & 53.57 & \underline{52.11} & 22.37 & 65.08 & 73.13 & 43.36 & 46.68 & 40.96 & 48.86 & 47.57 \\
\midrule
\textbf{Expert Merging++} & 31.14 & 56.53 & \textbf{52.35} & 22.37 & \underline{65.92} & \textbf{76.06} & \textbf{46.00} & \textbf{80.53} & \textbf{74.37} & \textbf{79.31} & \textbf{58.45} \\
\quad no chunking & \underline{31.16} & 56.77 & 47.71 & \underline{26.32} & 65.28 & 75.70 & \underline{45.31} & \underline{80.05} & \underline{73.85} & \underline{79.04} & \underline{58.11} \\
\quad chunk all layers (2x) & 31.03 & \underline{56.86} & 50.12 & 21.05 & 63.40 & 75.57 & 45.05 & 77.40 & 71.02 & 76.12 & 56.76 \\
\bottomrule
\end{tabular}%
}
\endgroup
\end{table}

\textit{InternVL2.5.} Table~\ref{tab:ablation-internvl} shows that, relative to \emph{Expert Merging}, removing hidden alignment slightly raises geometry (e.g., MathVista/MATH-Vision) but reduces grounding by 2–3 points, yielding a lower average—indicating hidden features help localization. Removing logit alignment severely hurts geometry and grounding (while sometimes inflating ChartQA), confirming logits are indispensable for cross-domain consistency. Eliminating coefficient regularization leads to a dramatic collapse, especially on grounding benchmarks. Guided chunking in \emph{Expert Merging++} improves OCR and grounding and achieves the best average, while chunking all layers uniformly harms geometry and grounding.

\begin{table}[t]
\centering
\caption{Ablation studies on Qwen2-VL. We report the impact of each loss component and chunking strategy.}
\label{tab:ablation-qwen2vl}
\begingroup
\setlength{\tabcolsep}{2.5pt}
\renewcommand{\arraystretch}{1.05}
\footnotesize
\resizebox{\linewidth}{!}{%
\begin{tabular}{l cc @ {\hspace{6pt}} cc @ {\hspace{6pt}} c @ {\hspace{6pt}} cc @ {\hspace{6pt}} ccc @ {\hspace{6pt}} c}
\toprule
 & \multicolumn{2}{c}{\textbf{VQA}} & \multicolumn{2}{c}{\textbf{Geometry}} & \multicolumn{1}{c}{\textbf{Chart}} & \multicolumn{2}{c}{\textbf{OCR}} & \multicolumn{3}{c}{\textbf{Grounding}} & \\
\cmidrule(lr){2-3} \cmidrule(lr){4-5} \cmidrule(lr){6-6} \cmidrule(lr){7-8} \cmidrule(lr){9-11}
\textbf{Variant} & \textbf{VizWiz} & \textbf{GQA} & \textbf{MathVista} & \textbf{MATH-Vision} & \textbf{ChartQA} & \textbf{TextVQA} & \textbf{OCRVQA} & \textbf{RefCOCO} & \textbf{RefCOCO+} & \textbf{RefCOCOg} & \textbf{Avg.} \\
\midrule
\textbf{Expert Merging} & 41.54 & 60.51 & 48.80 & \underline{42.11} & \textbf{73.00} & 81.41 & \underline{60.19} & 80.64 & \underline{66.15} & \textbf{79.02} & \underline{63.34} \\
\quad w/o hidden alignment & 41.64 & 60.35 & 47.66 & 34.21 & \underline{72.84} & \underline{81.59} & \textbf{60.38} & \underline{80.68} & 66.05 & 78.87 & 62.43 \\
\quad w/o logit alignment & \textbf{41.80} & 60.38 & 47.82 & 38.16 & 72.04 & 81.44 & 60.16 & 80.64 & 65.89 & 78.76 & 62.71 \\
\quad w/o task trade-offs & 41.56 & 60.53 & 48.38 & 36.84 & 72.80 & 81.49 & 59.96 & 80.51 & 66.00 & 78.82 & 62.69 \\
\quad w/o coefficient regularization & \underline{41.67} & \textbf{60.82} & \textbf{49.63} & 36.84 & 72.20 & \underline{81.59} & 60.12 & 78.33 & 63.90 & 76.59 & 62.17 \\
\midrule
\textbf{Expert Merging++} & 41.49 & 60.42 & \underline{49.33} & \textbf{44.74} & 72.60 & \textbf{81.65} & \underline{60.19} & \textbf{80.71} & \textbf{66.21} & \underline{79.00} & \textbf{63.63} \\
\quad no chunking & 41.54 & 60.51 & 48.80 & \underline{42.11} & \textbf{73.00} & 81.41 & \underline{60.19} & 80.64 & \underline{66.15} & \textbf{79.02} & \underline{63.34} \\
\quad chunk all layers (2x) & 41.47 & \underline{60.55} & 46.41 & 38.16 & 72.72 & 81.45 & 60.12 & 80.49 & 65.87 & 78.90 & 62.61 \\
\bottomrule
\end{tabular}%
}
\endgroup
\end{table}

\noindent\textit{Qwen2-VL.} Table~\ref{tab:ablation-qwen2vl} shows that sensitivity is milder: removing hidden or logit alignment changes the average only modestly, though MATH-Vision degrades without logits. Dropping regularization shifts scores unevenly (e.g., slight gains on GQA/MathVista but clear declines in grounding), underscoring the stability–variance trade-off. Importance-guided chunking improves MATH-Vision and the macro average; indiscriminate chunk-all-layers again underperforms.

\noindent\textit{Mistral-7B.} Table~\ref{tab:ablation-mistral} shows that removing logit alignment yields the largest drop among ablations, impacting both instruction following and code. The coefficient regularizer is likewise crucial: without it, average performance falls sharply. \emph{Expert Merging++} delivers the best average, with gains dominated by a large improvement on HumanEval and consistent lifts on GSM8K/MBPP, while AlpacaEval remains strong. Uniform chunking degrades results relative to targeted chunking.

\begin{table}[t]
\centering
\caption{Ablation studies on Mistral. We evaluate the effect of each loss component and chunking choice on text benchmarks.}
\label{tab:ablation-mistral}
\begingroup
\resizebox{0.75\linewidth}{!}{
\setlength{\tabcolsep}{5pt}
\renewcommand{\arraystretch}{1.05}
\footnotesize
\begin{tabular}{l c cc cc c}
\toprule
 & \multicolumn{1}{c}{\textbf{Chat}} & \multicolumn{2}{c}{\textbf{MATH}} & \multicolumn{2}{c}{\textbf{Code}} & \\
\cmidrule(lr){2-2} \cmidrule(lr){3-4} \cmidrule(lr){5-6}
\textbf{Variant} & \textbf{AlpacaEval} & \textbf{GSM8K} & \textbf{MATH} & \textbf{HumanEval} & \textbf{MBPP} & \textbf{Avg.} \\
\midrule
\textbf{Expert Merging} & 77.84 & \underline{58.68} & 13.88 & \underline{46.34} & 40.40 & \underline{47.43} \\
\quad w/o hidden alignment & \underline{77.99} & 57.09 & \underline{13.92} & 45.12 & \underline{41.40} & 47.10 \\
\quad w/o logit alignment & 69.68 & 56.10 & 13.32 & 38.41 & 40.60 & 43.62 \\
\quad w/o task trade-offs & \textbf{78.95} & 54.81 & 13.50 & 45.12 & \textbf{41.60} & 46.80 \\
\quad w/o coefficient regularization & 65.32 & 52.01 & 12.08 & 39.02 & 35.60 & 40.81 \\
\midrule
\textbf{Expert Merging++} & 75.51 & \textbf{59.44} & \textbf{13.94} & \textbf{53.66} & 41.00 & \textbf{48.71} \\
\quad no chunking & 77.84 & \underline{58.68} & 13.88 & \underline{46.34} & 40.40 & \underline{47.43} \\
\quad chunk all layers (2x) & 77.84 & 56.79 & 13.74 & 44.51 & 41.20 & 46.82 \\
\bottomrule
\end{tabular}
}
\endgroup
\end{table}

\subsection{Regularization Sensitivity}
\label{app:regu-sensitivity}

We evaluate the effect of the coefficient regularization strength on \emph{Expert Merging}, keeping all other settings fixed. Metrics and datasets match the main ablations.

\paragraph{Summary.}
Across backbones, moderate coefficient regularization yields the best macro averages: InternVL2.5 prefers a stronger setting ($\gamma\approx5$), while Qwen2-VL and Mistral-7B favor smaller values ($\gamma\approx0.8$). Gains come primarily from grounding (MLLMs) and code (LLMs), with mild trade-offs on Chart/geometry at very large $\gamma$. This behavior matches the role of the coefficient regularizer in curbing interference and preventing over-allocation. See Tables~\ref{tab:ablation-regu-internvl}, \ref{tab:ablation-regu-qwen2vl}, and \ref{tab:ablation-regu-mistral}.

\begin{table}[h]
\centering
\caption{Regularization sensitivity on InternVL2.5 for Expert Merging. Rows vary the coefficient regularization strength.}
\label{tab:ablation-regu-internvl}
\begingroup
\setlength{\tabcolsep}{2.5pt}
\renewcommand{\arraystretch}{1.05}
\footnotesize
\resizebox{\linewidth}{!}{%
\begin{tabular}{l cc @ {\hspace{6pt}} cc @ {\hspace{6pt}} c @ {\hspace{6pt}} cc @ {\hspace{6pt}} ccc @ {\hspace{6pt}} c}
\toprule
 & \multicolumn{2}{c}{\textbf{VQA}} & \multicolumn{2}{c}{\textbf{Geometry}} & \multicolumn{1}{c}{\textbf{Chart}} & \multicolumn{2}{c}{\textbf{OCR}} & \multicolumn{3}{c}{\textbf{Grounding}} & \\
\cmidrule(lr){2-3} \cmidrule(lr){4-5} \cmidrule(lr){6-6} \cmidrule(lr){7-8} \cmidrule(lr){9-11}
\boldmath$\gamma$ & \textbf{VizWiz} & \textbf{GQA} & \textbf{MathVista} & \textbf{MATH-Vision} & \textbf{ChartQA} & \textbf{TextVQA} & \textbf{OCRVQA} & \textbf{RefCOCO} & \textbf{RefCOCO+} & \textbf{RefCOCOg} & \textbf{Avg.} \\
\midrule
        \textbf{$\gamma{=}0$} & 29.15 & 51.06 & \textbf{51.93} & 19.74 & 65.64 & 72.52 & 42.42 & 32.46 & 29.06 & 40.97 & 43.49 \\ 
        \textbf{$\gamma{=}0.2$} & 29.75 & 52.97 & \underline{51.04} & 21.05 & 65.96 & 73.47 & 43.39 & 36.85 & 32.28 & 41.22 & 44.79 \\ 
        \textbf{$\gamma{=}0.4$} & 30.16 & 54.41 & 50.88 & 22.37 & \underline{66.44} & 73.94 & 44.08 & 38.02 & 32.85 & 38.54 & 45.16 \\ 
        \textbf{$\gamma{=}0.6$} & 30.49 & 55.12 & 49.47 & 22.37 & \textbf{66.76} & 74.43 & 44.69 & 40.26 & 34.15 & 37.03 & 45.47 \\ 
        \textbf{$\gamma{=}0.8$} & 30.63 & 55.52 & 49.98 & \textbf{26.32} & 66.16 & 74.87 & 45.05 & 44.13 & 38.03 & 39.97 & 47.06 \\ 
        \textbf{$\gamma{=}0.9$} & 30.83 & 55.72 & 48.19 & \underline{25.00} & 65.76 & 74.92 & 45.38 & 46.68 & 40.46 & 41.85 & 47.47 \\ 
        \textbf{$\gamma{=}1.0$} & 30.96 & 55.92 & 47.77 & 21.05 & 65.24 & 75.04 & 45.38 & 49.32 & 43.21 & 44.83 & 47.87 \\ 
        \textbf{$\gamma{=}1.5$} & 31.15 & 56.12 & 46.45 & 21.05 & 65.08 & 75.42 & 45.05 & 62.61 & 56.18 & 57.41 & 51.65 \\ 
        \textbf{$\gamma{=}2.0$} & \underline{31.23} & 56.27 & 47.73 & 21.05 & 64.16 & 75.76 & 44.95 & 71.77 & 65.62 & 68.14 & 54.66 \\ 
        $\gamma{=}3.0$ & \textbf{31.26} & 56.29 & 49.12 & 18.42 & 62.64 & \textbf{75.89} & 45.31 & 77.97 & 71.67 & 75.74 & \underline{56.43} \\ 
        $\gamma{=}5.0$ & 31.05 & 56.49 & 47.71 & 19.74 & 60.12 & \underline{75.87} & 45.44 & 79.08 & \underline{72.91} & 77.70 & \textbf{56.61} \\ 
        $\gamma{=}8.0$ & 30.94 & \underline{56.70} & 47.29 & 18.42 & 58.44 & 75.74 & \textbf{45.64} & \textbf{79.63} & \textbf{73.42} & \underline{78.00} & 56.42 \\ 
        $\gamma{=}10.0$ & 31.07 & \textbf{56.85} & 43.96 & 17.11 & 58.24 & 75.53 & \underline{45.57} & \underline{79.50} & \textbf{73.42} & \textbf{78.08} & 55.93 \\ 
\bottomrule
\end{tabular}%
}
\endgroup
\end{table}

\begin{table}[h]
\centering
\caption{Regularization sensitivity on Qwen2-VL for Expert Merging. Rows vary the coefficient regularization strength.}
\label{tab:ablation-regu-qwen2vl}
\begingroup
\setlength{\tabcolsep}{2.5pt}
\renewcommand{\arraystretch}{1.05}
\footnotesize
\resizebox{\linewidth}{!}{%
\begin{tabular}{l cc @ {\hspace{6pt}} cc @ {\hspace{6pt}} c @ {\hspace{6pt}} cc @ {\hspace{6pt}} ccc @ {\hspace{6pt}} c}
\toprule
 & \multicolumn{2}{c}{\textbf{VQA}} & \multicolumn{2}{c}{\textbf{Geometry}} & \multicolumn{1}{c}{\textbf{Chart}} & \multicolumn{2}{c}{\textbf{OCR}} & \multicolumn{3}{c}{\textbf{Grounding}} & \\
\cmidrule(lr){2-3} \cmidrule(lr){4-5} \cmidrule(lr){6-6} \cmidrule(lr){7-8} \cmidrule(lr){9-11}
\boldmath$\gamma$ & \textbf{VizWiz} & \textbf{GQA} & \textbf{MathVista} & \textbf{MATH-Vision} & \textbf{ChartQA} & \textbf{TextVQA} & \textbf{OCRVQA} & \textbf{RefCOCO} & \textbf{RefCOCO+} & \textbf{RefCOCOg} & \textbf{Avg.} \\
\midrule
        \textbf{$\gamma{=}0$} & 41.48 & \underline{60.51} & 48.80 & \underline{39.47} & 73.24 & 81.39 & 59.99 & \underline{80.59} & 66.17 & 78.93 & 63.06 \\ 
        \textbf{$\gamma{=}0.2$} & 41.45 & 60.42 & 47.96 & \underline{39.47} & \textbf{73.36} & 81.40 & 59.99 & \underline{80.59} & 66.06 & \underline{79.01} & 62.97 \\ 
        \textbf{$\gamma{=}0.4$} & 41.39 & 60.47 & \textbf{49.22} & \underline{39.47} & 73.20 & 81.39 & 60.06 & 80.55 & 66.13 & 78.91 & 63.08 \\ 
        \textbf{$\gamma{=}0.6$} & 41.43 & \textbf{60.52} & \textbf{49.22} & \underline{39.47} & \underline{73.32} & 81.34 & 60.06 & 80.57 & \textbf{66.24} & 78.94 & \underline{63.11} \\ 
        \textbf{$\gamma{=}0.8$} & \textbf{41.54} & \underline{60.51} & 48.80 & \textbf{42.11} & 73.00 & \underline{81.41} & \textbf{60.19} & \textbf{80.64} & 66.15 & \textbf{79.02} & \textbf{63.34} \\
        \textbf{$\gamma{=}0.9$} & 41.39 & 60.47 & 48.38 & \underline{39.47} & 73.12 & 81.32 & \underline{60.16} & 80.56 & 66.16 & 78.91 & 62.99 \\ 
        \textbf{$\gamma{=}1.0$} & 41.42 & 60.49 & 47.54 & \underline{39.47} & 73.28 & 81.40 & 59.93 & \underline{80.59} & 66.14 & 78.94 & 62.92 \\ 
        \textbf{$\gamma{=}1.5$} & 41.44 & 60.49 & \textbf{49.22} & \underline{39.47} & 73.28 & 81.36 & 60.06 & 80.53 & 66.15 & 78.92 & 63.09 \\ 
        \textbf{$\gamma{=}2.0$} & 41.42 & 60.44 & 48.38 & 38.16 & 73.12 & 81.39 & 60.06 & 80.58 & \underline{66.23} & 78.89 & 62.87 \\ 
        \textbf{$\gamma{=}3.0$} & \underline{41.52} & 60.43 & 48.38 & 38.16 & 73.28 & 81.35 & 60.09 & 80.57 & 66.16 & 78.98 & 62.89 \\
        \textbf{$\gamma{=}5.0$} & 41.50 & 60.43 & 47.96 & 38.16 & 73.16 & 81.39 & 60.06 & 80.55 & 66.16 & 78.96 & 62.83 \\
        \textbf{$\gamma{=}8.0$} & 41.46 & 60.45 & 48.80 & \underline{39.47} & 72.96 & 81.39 & 59.96 & 80.57 & 66.15 & 78.92 & 63.01 \\
        \textbf{$\gamma{=}10.0$} & 41.38 & 60.46 & \underline{48.92} & 38.16 & \textbf{73.36} & \textbf{81.44} & 59.93 & 80.56 & 66.13 & 78.96 & 62.93 \\
\bottomrule
\end{tabular}%
}
\endgroup
\end{table}

\begin{table}[h]
\centering
\caption{Regularization sensitivity on Mistral for Expert Merging. Rows vary the coefficient regularization strength.}
\label{tab:ablation-regu-mistral}
\begingroup
\resizebox{0.6\linewidth}{!}{
\setlength{\tabcolsep}{5pt}
\renewcommand{\arraystretch}{1.05}
\footnotesize
\begin{tabular}{l c cc cc c}
\toprule
 & \multicolumn{1}{c}{\textbf{Chat}} & \multicolumn{2}{c}{\textbf{MATH}} & \multicolumn{2}{c}{\textbf{Code}} & \\
\cmidrule(lr){2-2} \cmidrule(lr){3-4} \cmidrule(lr){5-6}
\boldmath$\gamma$ & \textbf{AlpacaEval} & \textbf{GSM8K} & \textbf{MATH} & \textbf{HumanEval} & \textbf{MBPP} & \textbf{Avg.} \\
\midrule
   \textbf{$\gamma{=}0$} & 65.32 & 52.01 & 12.08 & 39.02 & 35.80 & 40.85 \\ 
        \textbf{$\gamma{=}0.2$} & 75.53 & 55.42 & 13.58 & 43.90 & 39.40 & 45.57 \\ 
        \textbf{$\gamma{=}0.4$} & 75.61 & \underline{57.32} & \textbf{14.02} & 45.73 & 41.00 & 46.74 \\ 
        \textbf{$\gamma{=}0.6$} & 77.52 & \textbf{58.68} & 13.84 & \underline{46.95} & 39.40 & \underline{47.28} \\ 
        \textbf{$\gamma{=}0.8$} & 77.84 & \textbf{58.68} & \underline{13.88} & 46.34 & 40.40 & \textbf{47.43} \\ 
        \textbf{$\gamma{=}0.9$} & 76.66 & 56.71 & 13.22 & \textbf{48.78} & 41.00 & 47.27 \\ 
        \textbf{$\gamma{=}1.0$} & 78.78 & 57.01 & 13.60 & 43.29 & 39.00 & 46.34 \\ 
        \textbf{$\gamma{=}1.5$} & 79.01 & 56.25 & 12.24 & 45.73 & \underline{42.60} & 47.17 \\ 
        \textbf{$\gamma{=}2.0$} & \underline{79.28} & 56.41 & 12.00 & 44.51 & \underline{42.60} & 46.96 \\
        \textbf{$\gamma{=}3.0$} & \textbf{79.64} & 56.18 & 13.80 & 42.68 & \textbf{43.00} & 47.06 \\ 
        \textbf{$\gamma{=}5.0$} & 79.09 & 55.34 & 13.20 & 43.90 & 42.40 & 46.79 \\ 
        \textbf{$\gamma{=}8.0$} & 79.00 & 55.65 & 13.66 & 43.29 & 40.60 & 46.44 \\ 
        \textbf{$\gamma{=}10.0$} & 78.77 & 56.18 & 13.38 & 43.90 & 40.60 & 46.57 \\ 
\bottomrule
\end{tabular}
}
\endgroup
\end{table}

\paragraph{Backbone-specific results.}
For InternVL2.5, Table~\ref{tab:ablation-regu-internvl} shows the average rises from 43.49 at $\gamma{=}0$ to 56.61 at $\gamma{=}5.0$ (best), then plateaus (56.42 at $\gamma{=}8.0$; 55.93 at $\gamma{=}10.0$). The gain is driven largely by grounding: RefCOCO/RefCOCO+/RefCOCOg improve from 32.46/29.06/40.97 to 79.08/72.91/77.70 at $\gamma{=}5.0$ (79.63/73.42/78.00 at $\gamma{=}8.0$). ChartQA and MATH-Vision favor smaller $\gamma$ (66.76 at 0.6; 26.32 at 0.8), while VQA/OCR change little. Overall, $\gamma\approx5$ is the best trade-off.

For Qwen2-VL, Table~\ref{tab:ablation-regu-qwen2vl} shows sensitivity is mild: the average peaks at 63.34 for $\gamma{=}0.8$ and remains within about 0.5 over $\gamma\in[0,10]$. MATH-Vision benefits from moderate regularization (42.11 at 0.8 vs. 39.47 at 0), whereas VQA/OCR are nearly unchanged and grounding remains stable around 80.6/66.2/79.0. A smaller default ($\gamma\approx0.8$) is appropriate.

For Mistral-7B, Table~\ref{tab:ablation-regu-mistral} shows the average improves from 40.85 at $\gamma{=}0$ to 47.43 at $\gamma{=}0.8$ (best), then drifts slightly (47.27 at 0.9; 46.34 at 1.0). Task-wise, HumanEval peaks at 48.78 (0.9), MBPP at 43.00 (3.0), and AlpacaEval increases up to 79.64 (3.0). A default of $\gamma\approx0.8$ gives the strongest overall average.

\subsection{Sample Efficiency}
\label{app:sample-efficiency}

We report the impact of the number of unlabeled calibration samples used by \emph{Expert Merging}. Metrics follow the same protocol as above.

\paragraph{Summary.}
Across all backbones, performance saturates with very few unlabeled samples: best averages occur at $N\in\{5,10\}$, and larger $N$ brings marginal or negative returns. This indicates that layer-wise coefficient learning has low sample complexity, consistent with the small parameter count and the stabilizing effect of the regularizer and coefficient budget. See Tables~\ref{tab:ablation-sample-internvl}, \ref{tab:ablation-sample-qwen2vl}, and \ref{tab:ablation-sample-mistral}.

\begin{table}[h]
\centering
\caption{Sample efficiency on InternVL2.5 for Expert Merging. Rows vary the number of unlabeled calibration samples.}
\label{tab:ablation-sample-internvl}
\begingroup
\setlength{\tabcolsep}{2.5pt}
\renewcommand{\arraystretch}{1.05}
\footnotesize
\resizebox{\linewidth}{!}{%
\begin{tabular}{l cc @ {\hspace{6pt}} cc @ {\hspace{6pt}} c @ {\hspace{6pt}} cc @ {\hspace{6pt}} ccc @ {\hspace{6pt}} c}
\toprule
 & \multicolumn{2}{c}{\textbf{VQA}} & \multicolumn{2}{c}{\textbf{Geometry}} & \multicolumn{1}{c}{\textbf{Chart}} & \multicolumn{2}{c}{\textbf{OCR}} & \multicolumn{3}{c}{\textbf{Grounding}} & \\
\cmidrule(lr){2-3} \cmidrule(lr){4-5} \cmidrule(lr){6-6} \cmidrule(lr){7-8} \cmidrule(lr){9-11}
\boldmath$N$ & \textbf{VizWiz} & \textbf{GQA} & \textbf{MathVista} & \textbf{MATH-Vision} & \textbf{ChartQA} & \textbf{TextVQA} & \textbf{OCRVQA} & \textbf{RefCOCO} & \textbf{RefCOCO+} & \textbf{RefCOCOg} & \textbf{Avg.} \\
\midrule
        \textbf{$N{=}3$} & \underline{31.10} & \underline{56.75} & 46.91 & \underline{25.00} & 60.04 & 75.37 & \underline{45.44} & 77.41 & 70.75 & 75.86 & 56.46 \\ 
        \textbf{$N{=}5$} & \textbf{31.16} & \textbf{56.77} & 47.71 & \textbf{26.32} & \textbf{65.28} & \underline{75.70} & 45.31 & \textbf{80.05} & \textbf{73.85} & \textbf{79.04} & \textbf{58.11} \\ 
        \textbf{$N{=}10$} & 31.01 & 56.61 & \textbf{51.61} & 21.05 & 59.96 & 75.68 & \textbf{45.93} & \underline{79.76} & \underline{73.61} & \underline{78.33} & \underline{57.35} \\ 
        \textbf{$N{=}15$} & 30.93 & 56.57 & 50.18 & 22.37 & 60.84 & 75.38 & 44.95 & 78.84 & 72.68 & 77.08 & 56.98 \\ 
        \textbf{$N{=}20$} & 31.05 & 56.63 & 50.06 & 21.05 & \underline{62.04} & \textbf{75.74} & 44.73 & 79.20 & 72.45 & 76.61 & 56.95 \\ 
        \textbf{$N{=}30$} & 30.89 & 56.63 & 49.92 & \underline{25.00} & 59.80 & 75.39 & 44.79 & 78.48 & 72.10 & 76.23 & 56.92 \\ 
        \textbf{$N{=}50$} & 30.71 & 56.57 & \underline{50.34} & 21.05 & 56.28 & 75.22 & 45.28 & 78.38 & 71.75 & 76.39 & 56.19 \\ 
        \textbf{$N{=}100$} & 31.00 & 56.71 & 48.25 & 21.05 & 60.32 & 75.28 & 44.53 & 78.84 & 72.06 & 76.53 & 56.45 \\ 
\bottomrule
\end{tabular}%
}
\endgroup
\end{table}

\begin{table}[h]
\centering
\caption{Sample efficiency on Qwen2-VL for Expert Merging. Rows vary the number of unlabeled calibration samples.}
\label{tab:ablation-sample-qwen2vl}
\begingroup
\setlength{\tabcolsep}{2.5pt}
\renewcommand{\arraystretch}{1.05}
\footnotesize
\resizebox{\linewidth}{!}{%
\begin{tabular}{l cc @ {\hspace{6pt}} cc @ {\hspace{6pt}} c @ {\hspace{6pt}} cc @ {\hspace{6pt}} ccc @ {\hspace{6pt}} c}
\toprule
 & \multicolumn{2}{c}{\textbf{VQA}} & \multicolumn{2}{c}{\textbf{Geometry}} & \multicolumn{1}{c}{\textbf{Chart}} & \multicolumn{2}{c}{\textbf{OCR}} & \multicolumn{3}{c}{\textbf{Grounding}} & \\
\cmidrule(lr){2-3} \cmidrule(lr){4-5} \cmidrule(lr){6-6} \cmidrule(lr){7-8} \cmidrule(lr){9-11}
\boldmath$N$ & \textbf{VizWiz} & \textbf{GQA} & \textbf{MathVista} & \textbf{MATH-Vision} & \textbf{ChartQA} & \textbf{TextVQA} & \textbf{OCRVQA} & \textbf{RefCOCO} & \textbf{RefCOCO+} & \textbf{RefCOCOg} & \textbf{Avg.} \\
\midrule
        \textbf{$N{=}3$} & 41.42 & \textbf{60.53} & 48.80 & \underline{39.47} & \textbf{74.08} & 81.21 & 59.93 & 80.41 & 65.63 & 78.58 & 63.01 \\ 
        \textbf{$N{=}5$} & \underline{41.71} & 60.41 & 47.40 & \underline{39.47} & \underline{73.36} & 81.52 & \textbf{60.45} & \textbf{80.72} & \textbf{66.35} & \underline{79.02} & \underline{63.04} \\ 
        \textbf{$N{=}10$} & 41.54 & \underline{60.51} & 48.80 & \textbf{42.11} & 73.00 & 81.41 & 60.19 & 80.64 & \underline{66.15} & \underline{79.02} & \textbf{63.34} \\ 
        \textbf{$N{=}15$} & \textbf{41.75} & 60.34 & 49.75 & 34.21 & 71.84 & \textbf{81.66} & 60.29 & 80.60 & 66.17 & 78.93 & 62.55 \\ 
        \textbf{$N{=}20$} & 41.57 & 60.36 & \textbf{50.05} & 35.53 & 72.84 & 81.55 & 60.09 & 80.63 & 66.15 & \underline{79.00} & 62.78 \\ 
        \textbf{$N{=}30$} & 41.60 & 60.31 & \underline{49.93} & 32.89 & 73.00 & \underline{81.60} & 60.12 & 80.55 & 65.99 & 78.87 & 62.49 \\ 
        {$N{=}50$} & 41.60 & 60.38 & 48.92 & 35.53 & 72.80 & 81.58 & 60.35 & \underline{80.67} & 66.28 & \textbf{79.03} & 62.71 \\ 
        {$N{=}100$} & 41.53 & 60.44 & 49.22 & 36.84 & 73.12 & 81.55 & \underline{60.38} & \underline{80.69} & \underline{66.31} & 78.96 & 62.90 \\ 
\bottomrule
\end{tabular}%
}
\endgroup
\end{table}

\begin{table}[h]
\centering
\caption{Sample efficiency on Mistral for Expert Merging. Rows vary the number of unlabeled calibration samples.}
\label{tab:ablation-sample-mistral}
\begingroup
\resizebox{0.6\linewidth}{!}{
\setlength{\tabcolsep}{5pt}
\renewcommand{\arraystretch}{1.05}
\footnotesize
\begin{tabular}{l c cc cc c}
\toprule
 & \multicolumn{1}{c}{\textbf{Chat}} & \multicolumn{2}{c}{\textbf{MATH}} & \multicolumn{2}{c}{\textbf{Code}} & \\
\cmidrule(lr){2-2} \cmidrule(lr){3-4} \cmidrule(lr){5-6}
\boldmath$N$ & \textbf{AlpacaEval} & \textbf{GSM8K} & \textbf{MATH} & \textbf{HumanEval} & \textbf{MBPP} & \textbf{Avg.} \\
\midrule
        \textbf{$N{=}3$} & 75.41 & \underline{58.07} & 13.80 & \textbf{48.17} & \underline{40.40} & \underline{47.17} \\ 
        \textbf{$N{=}5$} & \underline{77.84} & \textbf{58.68} & \underline{13.88} & 46.34 & \underline{40.40} & \textbf{47.43} \\ 
        \textbf{$N{=}10$} & 75.69 & 56.71 & 13.66 & 45.12 & \textbf{40.60} & 46.36 \\ 
        \textbf{$N{=}15$} & 76.22 & 57.24 & \textbf{14.30} & 45.12 & 38.00 & 46.18 \\ 
        \textbf{$N{=}20$} & \textbf{77.88} & 57.47 & 13.42 & 41.46 & 39.00 & 45.85 \\ 
        \textbf{$N{=}30$} & 76.89 & 57.62 & 13.80 & 45.73 & 39.00 & 46.61 \\ 
        \textbf{$N{=}50$} & 76.65 & 57.85 & 13.82 & \underline{47.56} & 39.40 & 47.06 \\ 
        \textbf{$N{=}100$} & 77.23 & 57.54 & \underline{13.88} & 46.34 & 39.00 & 46.80 \\ 
\bottomrule
\end{tabular}
}
\endgroup
\end{table}

\paragraph{Backbone-specific results.}
For InternVL2.5, Table~\ref{tab:ablation-sample-internvl} shows the average is 56.46 at $N{=}3$, improves to 58.11 at $N{=}5$ (best), and is 57.35 at $N{=}10$. Grounding peaks at $N{=}5$ (RefCOCO/RefCOCO+/RefCOCOg: 80.05/73.85/79.04), while geometry varies (MathVista 51.61 at 10; MATH-Vision 26.32 at 5). Larger $N$ yields little benefit.

For Qwen2-VL, Table~\ref{tab:ablation-sample-qwen2vl} shows results plateau: the average is 63.01 at $N{=}3$, 63.04 at $N{=}5$, and 63.34 at $N{=}10$ (best). MATH-Vision benefits most from $N{=}10$ (42.11), while other domains change marginally (e.g., TextVQA 81.21--81.66). Beyond $N{=}10$, improvements are small or negative.

For Mistral-7B, Table~\ref{tab:ablation-sample-mistral} shows the average is 47.17 at $N{=}3$ and 47.43 at $N{=}5$ (best). Tasks prefer different regimes (HumanEval 48.17 at 3; MBPP 40.60 at 10), but overall a handful of samples per domain suffices.

\paragraph{Takeaway.}
Regularization and data efficiency work in tandem: modest $\gamma$ reduces cross-domain interference so that a small calibration set captures task signatures reliably, and importance-guided chunking (Section~\ref{subsec:expert-merging}) further channels capacity to later layers where gains concentrate (grounding/code task).

\end{document}